\documentclass[conference]{IEEEtran}
\IEEEoverridecommandlockouts
\usepackage{cite}
\usepackage[utf8]{}
\usepackage{amsmath,amssymb,amsfonts}
\usepackage{algorithmic}
\usepackage{graphicx}
\usepackage{textcomp}
\usepackage{tabularx}
\usepackage{xcolor}
\usepackage{subcaption}
\def\BibTeX{{\rm B\kern-.05em{\sc i\kern-.025em b}\kern-.08em
    T\kern-.1667em\lower.7ex\hbox{E}\kern-.125emX}}
\begin{document}

\title{Text-Conditioned Diffusion Model \\for High-Fidelity Korean Font Generation
}

\author{
    Abdul Sami, Avinash Kumar, Irfanullah Memon, Youngwon Jo, Muhammad Rizwan, Jaeyong Choi\\
    \IEEEauthorblockA{
        School of Computer Science and Engineering, Soongsil University, Seoul 06978, Korea\\
        Emails: \{sami, kumar, iumemon, ywjo, mrizwan\}@soongsil.ac.kr, choi@ssu.ac.kr
    }
}

\maketitle
\begin{abstract}
Automatic font generation (AFG) is the process of creating a new font using only a few examples of the style images. Generating fonts for complex languages like Korean and Chinese, particularly in handwritten styles, presents significant challenges. Traditional AFGs, like Generative adversarial networks (GANs) and Variational Auto-Encoders (VAEs), are usually unstable during training and often face mode collapse problems. They also struggle to capture fine details within font images. To address these problems, we present a diffusion-based AFG method which generates high-quality, diverse Korean font images using only a single reference image, focusing on handwritten and printed styles. Our approach refines noisy images incrementally, ensuring stable training and visually appealing results. A key innovation is our text encoder, which processes phonetic representations to generate accurate and contextually correct characters, even for unseen characters. We used a pre-trained style encoder from DG-FONT to effectively and accurately encode the style images. To further enhance the generation quality, we used perceptual loss that guides the model to focus on the global style of generated images.  Experimental results on over 2000 Korean characters demonstrate that our model consistently generates accurate and detailed font images and outperforms benchmark methods, making it a reliable tool for generating authentic Korean fonts across different styles.

\end{abstract}
\vspace{0.5em} %
\vspace{0.5em} %
\section{ \textbf{Introduction}}
Writing has arguably been one of the most significant means of human expression across cultures and languages. From ancient carvings in stone to the graceful strokes of calligraphy, each writing style tells a story and reflects the emotions of its time. The creation of Hangeul in the 15th century democratized writing for the first time in Korea and consequently laid the foundation for a tradition of beautiful handwritten characters. These Handwriting fonts based on this script share a common heritage of beauty and personal expression, and reproducing these font styles for over 11,000 Hangeul characters as digital fonts is very challenging. Moreover, Chinese and Japanese languages face similar issues with even more complex scripts\cite{yang2024fontdiffuser}. Creating new fonts is a labour intensive process 
 and requires a lot of effort and sometimes takes years to fully capture the true essence of text style.\\
\indent With the recent advances in machine learning, more and more avenues of font generation have been opened up, whereby tools can automatically generate a vast array of styles. Early Generative Adversarial Networks (GANs) based Automatic Font Generation (AFG) methods \cite{xie2021dg,liu2019few, hassan2023real,muhammad4370420learning} and Variational Auto-Encoders (VAEs) based AFG made significant strides in automating the font design process. However, these techniques were not without their challenges. One of the main issues with GANs was their notorious difficulty in training; they often required careful balancing between the generator and discriminator. This instability could make it hard to achieve high-quality font generation, leading to missing strokes, style inconsistency, random artifacts, blurriness, layout errors or inconsistencies in the generated characters\cite{dhariwal2021diffusion}. VAEs, on the other hand, often faced difficulties capturing the intricate details required for complex scripts.\\
\indent Both GANs and VAEs need large amounts of labeled data to train properly, which can be a significant barrier in languages with rich and complex character sets like Korean and Chinese. Even with sufficient data, these models cannot generate font images consistent in style throughout the entire character set and are correct content-wise. That became a big problem regarding scripts with more complex structures, where the stroke order and subtleties of shape are critical components of textual authenticity. Thus, while they provided some ground for the automated font generation that followed, they often needed to fully capture and preserve the details of the handwritten characters in languages with a more complex structure.\\
\indent Now, with the new diffusion models, we have a new and improved way to generate high-fidelity images\cite{dhariwal2021diffusion,ho2020denoising,choi2021conditioning,saharia2022palette,sohl2015deep,nichol2021improved,rombach2022high}. They address all the challenges faced by GANs and VAEs.\cite{he2024diff,yang2024fontdiffuser} Were the first to work on AFG using diffusion models. \cite{he2024diff} was the first such work in diffusion-based AFG, which added techniques to stabilize and resolve the mode collapse problem in GAN-based models. The recent work of \cite{yang2024fontdiffuser}  integrates multiscale content aggregation and style contrastive learning, boosting capacities to generate complex characters and managing significant variations within the style. However, even after all these works, it is often seen that there is much room for effort in diffusion-based font generation, in particular, the preservation of nuanced features and stylistic variation. Our work advances these ideas, developing and pushing diffusion model capabilities to explore new levels of automated font design, producing artistically diverse and accurate fonts.\\
\indent In this paper, we introduce DK-Font, a novel one-shot diffusion-based model for generating high-quality fonts using only one reference image. Our approach overcomes some of the problems of previous models, in particular, component misplacement, content loss, and inconsistent style. we introduce some novel improvements in font generation to increase the robustness and preciseness of the generated font images. The improvements include incorporating a text encoder that processes phonetic character representations to boost the model's content understanding and minimize errors. Moreover, our improved encoder is able to unify content, style, and stroke information, which makes it possible to generate complex structures of characters with higher accuracy. The use of perceptual loss during training further strengthens the fact that the generated characters are clean and whole, preserving both the structure and aesthetic integrity of the fonts. This additional contextual information further allows DK-Font to generate complete set of fonts using just a single reference image. The generated font images are not only visually similar to the original style but they also consistently maintain the intended character structure and style in all characters\\
\indent In summary, We introduce the methodology DK-Font and present its full-scale evaluation with over 2,000 Korean characters and benchmarking against the existing model. The results show that DK-Font is a significant step toward font generation, in general, and much more so for complex languages like Korean.

\section{\textbf{Related Work}}
\vspace{0.5em} %
\subsection{\textbf{Image-to-Image Translation}}
Image-to-image (I2I) translation is all about transforming an image from one type or style to another while keeping its essential content intact. One of the first approaches to this was Pix2Pix, which used a type of neural network called GANs to learn how to translate images, but it needed paired examples of source and target images, which isn’t always practical\cite{Isola_2017_CVPR}.\\
\indent To get around this, CycleGAN introduced a method that didn’t require paired images\cite{Zhu_2017_ICCV}. It made sure that if you translated an image to another style and then back again, you’d end up with something close to the original, which was a big step forward.\\
\indent Building on this, the UNIT framework combined GANs with another technique called VAEs to create a shared space where images from different styles could be more easily translated between each other. Other models like MUNIT and FUNIT went further by separating the content of an image from its style, allowing for more flexibility and variety in the results\cite{liu2019few}.\\
\indent More recently, diffusion models have been making waves in this area. Unlike GANs, which generate images directly, diffusion models work by gradually refining a noisy image until it becomes clear. Techniques like ILVR and Palette have shown that these models can produce high-quality images with better control over the output\cite{choi2021conditioning} \cite{saharia2022palette}.
However, generating fonts is a bit different and more challenging because it’s not just about changing the style but also about preserving the specific shapes and structures that define each letter.\\
\indent Most existing methods focus on visual changes like color or texture, which doesn’t fully address the needs of font design.
Our work aims to adapt these diffusion models specifically for font generation, tackling the unique challenges involved and pushing forward the automation of generating new fonts.  
\subsection{\textbf{Few-Shot Font Generation}}
Few-shot font generation aims to create a complete set of fonts using only a small number of reference images. This approach allows designers to rapidly generate new fonts by applying the style of a few characters to an entire alphabet or character set.\\
\indent Traditional methods for few-shot font generation often use image-to-image translation techniques, where the style from a few reference characters is transferred to others. Some approaches integrate font-specific knowledge, such as stroke details, to improve accuracy. For example, DG-Font uses deformable convolutional layers to capture style more effectively, while SC-Font incorporates stroke information to maintain structural correctness\cite{xie2021dg}.\\
\indent Despite these advancements, many methods rely on GANs, which can be unstable and produce inconsistent results. To address this, newer techniques have emerged that avoid using predefined font knowledge, aiming for greater flexibility. FS-Font explores spatial relationships between content and style, and CF-Font introduces an iterative process for refining style transfer. However, challenges remain in generating complex characters and handling diverse styles.
Our work seeks to overcome these issues, focusing on more robust and flexible approaches to few-shot font generation.

\subsection{\textbf{Diffusion Model }}
\begin{figure*}[h!]
    \centering
    \includegraphics[width=0.8\textwidth, height=0.2\textheight]{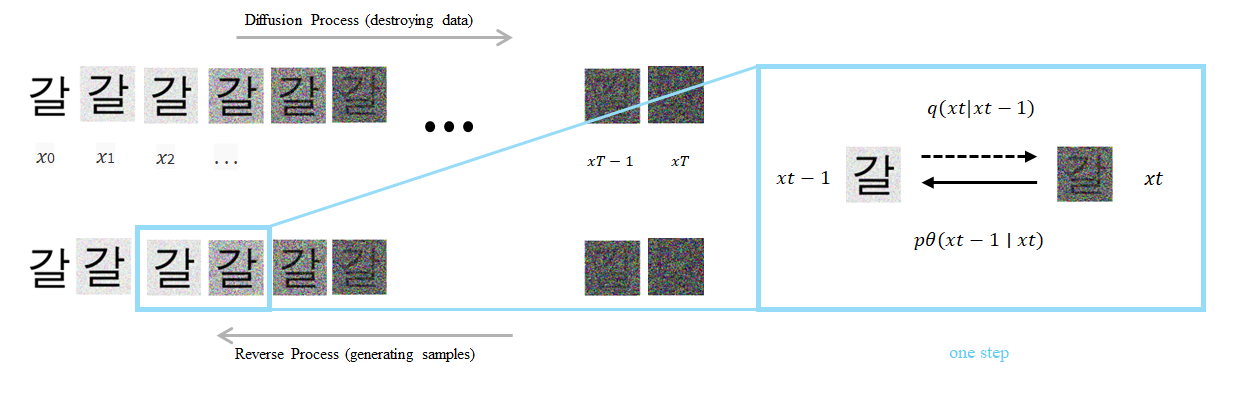}
    \caption{\textbf{Overview of the Diffusion Model Process.} The forward diffusion process (top row) transforms the original image into a noisy representation by gradually adding noise. The reverse process (bottom row) uses a deep learning model to progressively remove the noise, ultimately reconstructing the original data. The enlarged section on the right illustrates a single reverse step, where noise is estimated and removed from the current state \( x_t \) to approximate the previous state \( x_{t-1} \).}
    \label{fig:diffusion_model1}
\end{figure*}
\begin{figure*}[h!]
    \centering
    \includegraphics[width=0.8\textwidth, height=0.3\textheight]{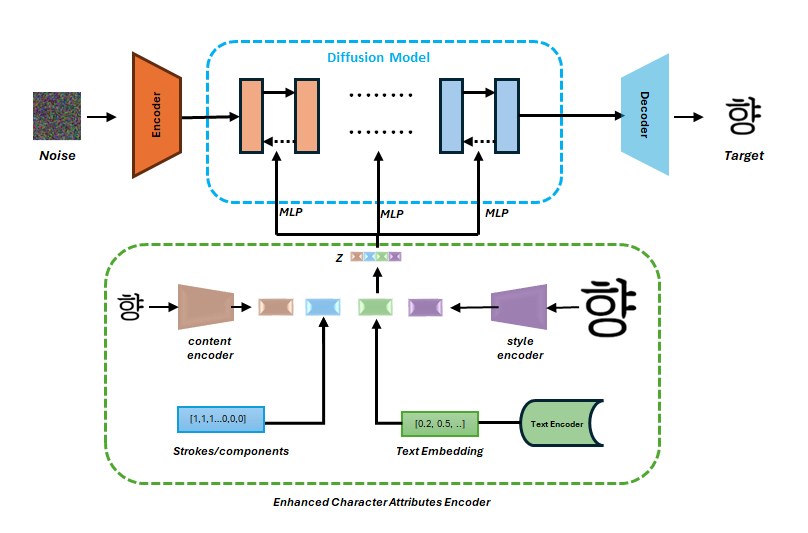}
    \caption{The framework of our model refines noisy input to generate font images. An Enhanced Character Attributes Encoder integrates strokes, components, and style features with text embeddings to guide this process. }
    \label{fig:diffusion_model}
\end{figure*}

Diffusion models have recently emerged as a powerful tool in generative modeling, providing an alternative to traditional methods like GANs and VAEs. These models work by iteratively adding noise to an image and then learning to reverse the process to generate high-quality samples. The concept was first introduced by Sohl-Dickstein Weiss, Maheswaranathan, and Ganguli\cite{sohl2015deep} and later refined by Ho, Jain, and Abbeel\cite{ho2020denoising} with the development of Denoising Diffusion Probabilistic Models (DDPMs), which have shown remarkable success in various image synthesis tasks. \\
\indent Building on these foundations, several advancements have been made to improve the efficiency and performance of diffusion models. Nichol and Dhariwal\cite{nichol2021improved} introduced modifications that reduce the number of steps required for sampling, making diffusion models more practical for real-world applications. Additionally, Rombach proposed Latent Diffusion Models (LDMs), which operate in a lower-dimensional latent space, further enhancing computational efficiency without compromising image quality\cite{rombach2022high}. \\ 
\indent In the field of font generation, diffusion models are still quite new but show a lot of potential. The diffusion-based AFG was first introduced by \cite{he2024diff}, which stabilized and resolved the mode collapse problem related to GAN-based models. The recent work of \cite{yang2024fontdiffuser} integrates multi-scale content aggregation and style contrastive learning, boosting capacities in generating complex characters and managing significant variations in style. However, even after all these works, it is often seen that there is much room for improvement in diffusion-based font generation, in particular, the preservation of nuanced features and stylistic variation.  \\ 
\indent Our work aims to build on these ideas, refining and expanding the capabilities of diffusion models to push the boundaries of automated font design, generating artistically diverse and technically precise fonts.
\section{\textbf{METHODOLOGY}} 
\vspace{0.5em} %
% \begin{figure*}[h!]
%     \centering
%     \includegraphics[width=0.9\textwidth, height=0.2\textheight]{fig.PNG}
%     \caption{The diffusion model first gradually adds noise to destroy the data, then carefully removes the noise step by step to generate new samples. The noise is estimated using a deep-learning model with set parameters. }
%     \label{fig:diffusion_model}
% \end{figure*}
In this section, we present our innovative approach to automated font generation using a diffusion model. Our primary goal is to generate fonts that are both high-quality and diverse in style while maintaining the intricate details and structural consistency crucial in font design.
\subsection{\textbf{Diffusion Process (Forward Process)}}
Our method begins with the forward pass, where we gradually introduce noise to the original font data, denoted as \(x_0\). This process generates a sequence of progressively noisier versions of the data, labeled as \(x_1, x_2, \dots, x_T\), where \(x_T\) represents the fully noisy sample as shown in Fig. \ref{fig:diffusion_model1}.\\

\indent Mathematically, the forward process is modelled as a Markov chain, where each step \(x_t\) is sampled from a Gaussian distribution conditioned on the previous step \(x_{t-1}\). Specifically, this is expressed as:
\begin{align}
    x_t \sim q(x_t \mid x_{t-1}) &= \mathcal{N}(x_t; \sqrt{1 - \beta_t} x_{t-1}, \beta_t I), \label{eqn1}
\end{align}
where, \(\beta_t\) is a predefined variance schedule that controls the amount of noise added at each step, and \(I\) is the identity matrix.
\subsection{\textbf{Reverse Process (Denoising Process)}}
After reaching the noisy sample \(x_T\), the reverse process begins. This process aims to gradually remove the noise, thereby recovering the original font data \(x_0\). The reverse process is also modelled as a Markov chain, where each step \(x_{t-1}\) is sampled from a distribution conditioned on the current step \(x_t\): see Fig.\ref{fig:diffusion_model1}.
\begin{align}
    x_{t-1} \sim p_\theta(x_{t-1} \mid x_t) &= \mathcal{N}(x_{t-1}; \mu_\theta(x_t, t), \Sigma_\theta(x_t, t)), \label{eqn2}
\end{align}
Here, \(\mu_\theta(x_t, t)\) and \(\Sigma_\theta(x_t, t)\) are the mean and covariance predicted by a neural network parameterized by \(\theta\), which learns to denoise the samples.

\subsection{\textbf{Gaussian Approximation and Training}}

If \( \beta_t \) is small enough, the forward process can be approximated as a Gaussian distribution directly conditioned on \( x_0 \):
\begin{align}
    q(x_t \mid x_0) &= \mathcal{N}(x_t; \sqrt{\alpha_t} x_0, (1 - \alpha_t) I), \label{eqn3}
\end{align}
where \( \alpha_t = \prod_{s=1}^{t}(1 - \beta_s) \).
\\
This leads to an approximation of the posterior distribution \( q(x_{t-1} \mid x_t, x_0) \) as:
\begin{align}
    q(x_{t-1} \mid x_t, x_0) &= \mathcal{N}(x_{t-1}; \tilde{\mu}_t(x_t, x_0), \tilde{\beta}_t I), \label{eqn4}
\end{align}
where the mean \( \tilde{\mu}_t(x_t, x_0) \), And variance \( \tilde{\beta}_t \) are given by:
\begin{align}
    \tilde{\mu}_t(x_t, x_0) &= \frac{\sqrt{\alpha_{t-1}} \beta_t}{1 - \alpha_t} x_0 + \frac{\sqrt{\alpha_t} (1 - \alpha_{t-1})}{1 - \alpha_t} x_t, \label{eqn5} \\
    \tilde{\beta}_t &= \frac{1 - \alpha_{t-1}}{1 - \alpha_t} \beta_t. \label{eqn6}
\end{align}

The training process involves teaching a neural network to predict the noise \( \epsilon_\theta(x_t) \) added to \( x_0 \). This is done by minimizing the difference between the predicted noise and the actual noise introduced during the forward process, based on the equation:
\begin{align}
    x_0 &= \frac{1}{\sqrt{\alpha_t}} x_t - \sqrt{\frac{\beta_t}{1 - \alpha_t}} \epsilon_\theta(x_t). \label{eqn7}
\end{align}

\subsection{\textbf{DK-Font: Enhancement and Innovations}}
Our work is inspired by the framework of Diff-Font while addressing key issues that we experienced. Firstly, character shuffling or mixing up the components of the character, leading to errors where one character might resemble another. Secondly, content loss where generated characters often appeared unclear or incomplete, especially with complex characters. The model had difficulties in maintaining the structural integrity of these complex characters, resulting in blurred or missing details. Thirdly, the model was unable to maintain a consistent style across all generated characters. This led to some characters not matching the intended style, resulting in an inconsistent appearance of the font. Lastly, despite the use of stroke guidelines, the model still made errors when generating characters with challenging, complex, or rare structures.
\\
\indent The proposed DK-Font presents a unique combination of text encoders and incorporates perceptual loss, so improving the precision and resilience of the font generation procedure. This model has been trained using an extensive dataset consisting of 2,350 Korean characters. Every individual character in this dataset is associated with its corresponding phonetic representation and stored in a standalone text file. The inclusion of this supplementary information enhances the model's comprehension of the content it must produce, so facilitating the development of characters that are more accurate and suitable for the given context.

We maintain the foundational U-Net architecture used in Diff-Font but introduce several key enhancements as shown in Fig. \ref{fig:diffusion_model}.\\
\indent\textbf{Text Encoder:} We introduce a text encoder to process the phonetic representations of the Korean characters. This encoder transforms the text input into a dense vector that captures the meaning and structure of the character. By including this textual information, the model better understands the characters, which helps reduce errors like character shuffling and content loss.\\
\indent\textbf{Enhanced Character Attributes Encoder:} The character attributes encoder now incorporates the outputs from the text encoder along with the usual content, style, and stroke information. These combined inputs form a comprehensive latent variable z that the diffusion model uses to generate the character images. This combination ensures that the model has multiple sources of information, improving its ability to accurately replicate complex character structures.\\
\indent\textbf{Perceptual Loss:} We used perceptual loss during training to enhance the quality of the generated characters and make the generated characters less ambiguous and much more complete. Contrary to a pixel-wise loss, the perceptual loss is designed to compare a high-level abstraction of the generated image and the target image with the help of a pre-trained convolutional neural network. We used VGG-19 as a feature extractor.  This approach helps to ensure that the generated images do not only look similar to the target images in the basic contour but also in finer details, which results in more accurate character generation.

\begin{figure*}[tbp]
    \centering
    \includegraphics[width=\linewidth]{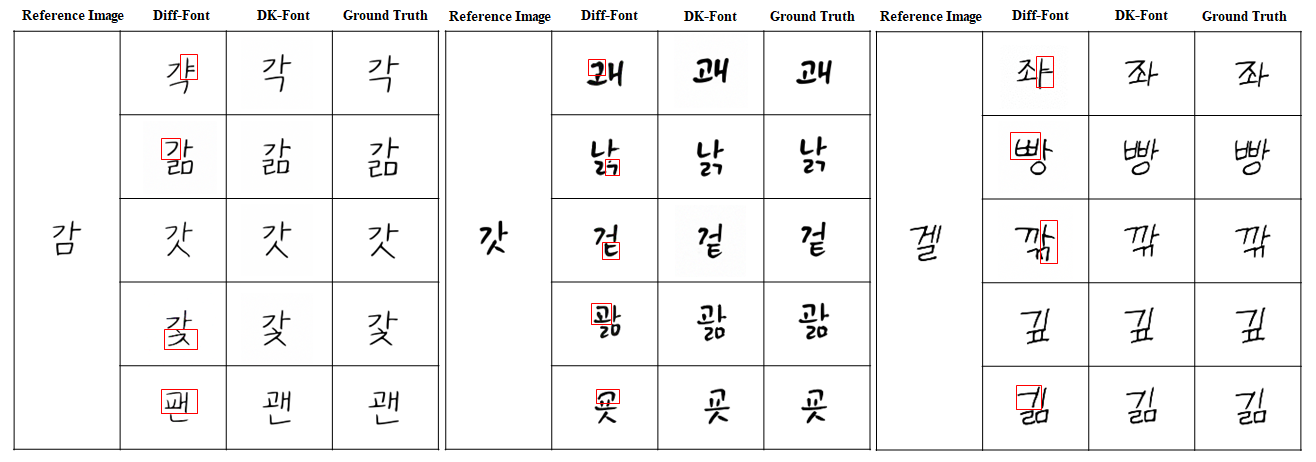}
    \caption{Comparison of our method with Diff-Font and ground truth on handwritten Korean fonts.}
    \label{fig:handwritten}
\end{figure*}

\begin{figure*}[tbp]
    \centering
    \includegraphics[width=\linewidth]{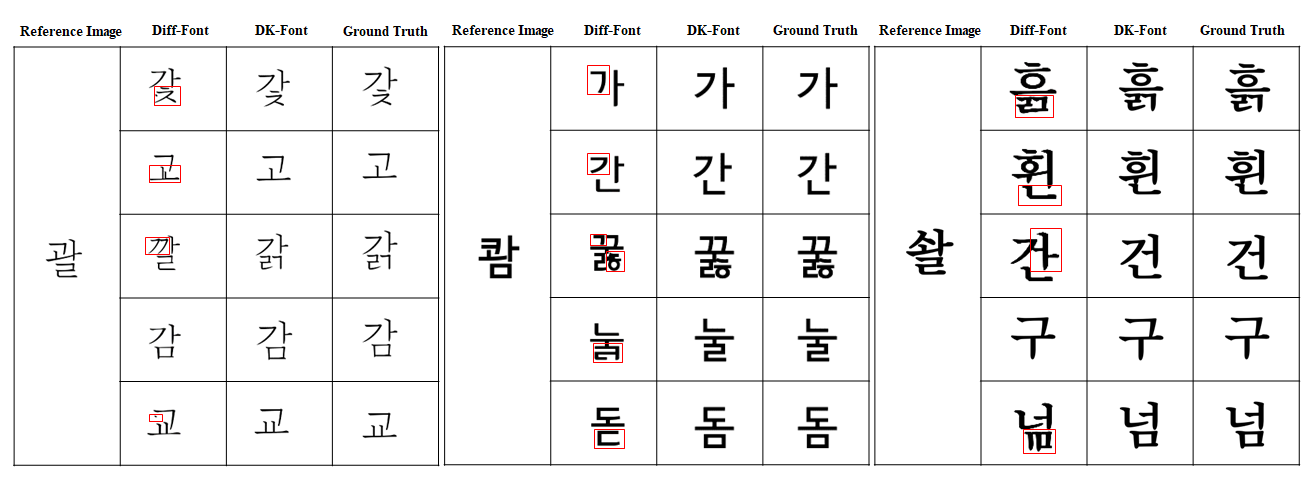}
    \caption{Comparison of our method with Diff-Font and ground truth on printed fonts.}
    \label{fig:printed}
\end{figure*}

\textbf{Training Process}:
We train the model using a combination of content, style, strokes, and the new text-based inputs. The diffusion process, guided by the U-Net architecture, gradually denoises the input images based on the latent variable z, which now includes the phonetic information. The training objective is a mix of Mean Squared Error (MSE) loss and perceptual loss, ensuring that the generated images are both structurally accurate and visually consistent compared to the target.
\vspace{0.5em} 
\section{\textbf{EXPERIMENTS}\\} 
\vspace{0.5em} 
\textbf{Dataset and Evaluation Metrics} \\
For our experiments, we curated two robust datasets specifically designed to evaluate the performance of our diffusion-based font generation model. These datasets focus exclusively on Korean (Hangeul) characters, comprising both printed and handwritten styles, providing a diverse and comprehensive foundation for training and testing.\\ 

\indent\textbf{Hangeul Dataset \\}
We assembled a collection of 210 font styles, comprising 100 handwritten and 100 printed fonts. Each font style contains 2,350 of the most commonly used Hangeul characters, ensuring comprehensive coverage of everyday language needs. For each category, we trained the model separately: 100 handwritten fonts were used for training, and 5 additional handwritten fonts were set aside for testing. These 5 handwritten fonts were not included in the training process, ensuring they are entirely unseen by the model. Similarly, 100 printed fonts were used for training, with 5 separate printed fonts reserved for testing, which the model did not encounter during training. This setup allows us to rigorously evaluate the model’s ability to generalize across both familiar and novel font styles within each category, as summarized in \text{TABLE I} 
\\
\begin{table}[h!]
\centering
\caption{Summary of Hangeul Dataset}
\label{table:font_styles}
\scalebox{1.2}{
\begin{tabular}{lcccc}
\hline
\textbf{Category} & \textbf{Characters} & \textbf{Fonts} & \textbf{Training} & \textbf{Testing} \\ \hline
Handwritten       & 2,350               & 105            & 100               & 5               \\ 
Printed           & 2,350               & 105            & 100               & 5               \\ \hline
\textbf{Total}    & 2,350               & 210            & 200               & 10              \\ \hline
\end{tabular}}
\end{table}

For all datasets, the images were standardized to a size of 128×128 pixels, which is optimal for our deep-learning models. \\
\indent For the handwritten fonts, Figure~\ref{fig:handwritten} illustrates a visual comparison of our method against Diff-Font and the ground truth. As can be seen, DK-Font produces high-quality results that closely resemble the original handwritten fonts, maintaining structural details and style consistency. Diff-Font, on the other hand, suffers from missing strokes and inconsistencies.\\

For printed fonts, the comparison in Figure~\ref{fig:printed} shows similar improvements. DK-Font effectively preserves the intricate details of printed characters and maintains consistency across the generated fonts. Diff-Font, however, often introduces artifacts and fails to maintain structural integrity.\\

\indent\textbf{Evaluation Metrics}
To evaluate the effectiveness of our model, we used several commonly used metrics in image generation:

\begin{itemize}
    \item SSIM (Structural Similarity Index): Assesses structural similarity between images, considering luminance, contrast, and overall structure to gauge how closely our generated fonts resemble the target images.
\end{itemize}

\begin{itemize}
    \item RMSE (Root Mean Square Error):  Measures pixel-by-pixel differences between generated and target images for a straightforward accuracy assessment.
\end{itemize}

\begin{itemize}
    \item LPIPS (Learned Perceptual Image Patch Similarity): Evaluates perceptual similarity using deep neural network features, aligning more closely with human visual perception.
\end{itemize}

\begin{itemize}
    \item FID (Fréchet Inception Distance):  FID Measures the distance between the generated images using and the distribution of the real images, providing insights into the quality and realism of the generated fonts.\\
\end{itemize} 

\textbf{Implementation Details} 

Our model was implemented using the AdamW optimizer, with parameters set to $\beta_1 = 0.9$ and $\beta_2 = 0.999$. All images in the dataset were resized to 128 $\times$ 128 pixels to ensure consistency and optimal processing by the model.
Due to limited computing power, we used a batch size of 16 during training to balance memory usage and computational efficiency. To prevent over-fitting and improve generalization of the model, we dropout with probability 0.1.
We employed a diffusion process with 1,000 steps. Initially, we experimented with a basic linear noise scheduler, but the results were not as promising. We switched to a cosine noise scheduler to improve performance, which provided better results due to its more advanced attributes for managing noise during the diffusion process.
The learning rate was fixed at 0.001 throughout the training process. We trained the model for 50,000 iterations, allowing sufficient time to learn and refine its ability to generate high-quality font images.
\\
\begin{table}[h!]
\centering
\caption{Quantitative Comparison of Different Methods}
\label{tab:quantitative_comparison}
\scalebox{1.2}{
\begin{tabular}{lcccc}
\hline
\textbf{Methods}  & \textbf{  SSIM} $\uparrow$  & \textbf{  RMSE} $\downarrow$ &  \textbf{  LPIPS} $\downarrow$  & \textbf{  FID} $\downarrow$ \\ \hline
Diff-Font       & 0.812          & 0.196         & 0.072          & 10.69        \\ \hline
\textbf{DK-Font} & \textbf{0.857} & \textbf{0.123} & \textbf{0.063} & \textbf{10.446} \\ \hline
\end{tabular}}
\end{table}

\textbf{Quantitative Comparison} \\
\indent In this section, we compare our DK-Font with Diff-Font. Only one reference character image with the target font is utilized during the generation process. To ensure a fair and comprehensive evaluation we selected a diverse set of test fonts that represent a broad range of styles.\\
\indent The summarized results are shown in \text{TABLE II}, from which it can be derived that the proposed method generally outperforms Diff-Font in terms of key metrics, including SSIM, RMSE, LPIPS, and FID. Most evaluations show that DK-Font consistently yields the best performance, demonstrating the strong potential of the proposed solution for generating high-quality fonts.\\

\section{\textbf{CONCLUSION}\\}
\vspace{0.5em} 
In this paper, we introduced DK-Font, a text-conditioned diffusion model for high-fidelity font generation method, with a special emphasis on the challenges of Korean font generation. Compared to the previous methods such as GANs and VAEs, our model is able to preserve the structure and style of each character and generate visually pleasing and diverse fonts. The use of text encoder and perceptual loss was effective in enhancing the quality of the generated fonts for unseen characters. The results of our experiments showed that DK-Font outperforms other methods in terms of the key metrics and can be used as an efficient tool for automated font generation.

\bibliographystyle{IEEEtran}  % Specifies the bibliography style (IEEEtran for IEEE)
\bibliography{conference_101719.bbl}     % Specifies the name of the .bib file without the extension

\begin{thebibliography}{10}
\providecommand{\url}[1]{#1}
\csname url@samestyle\endcsname
\providecommand{\newblock}{\relax}
\providecommand{\bibinfo}[2]{#2}
\providecommand{\BIBentrySTDinterwordspacing}{\spaceskip=0pt\relax}
\providecommand{\BIBentryALTinterwordstretchfactor}{4}
\providecommand{\BIBentryALTinterwordspacing}{\spaceskip=\fontdimen2\font plus
\BIBentryALTinterwordstretchfactor\fontdimen3\font minus \fontdimen4\font\relax}
\providecommand{\BIBforeignlanguage}[2]{{%
\expandafter\ifx\csname l@#1\endcsname\relax
\typeout{** WARNING: IEEEtran.bst: No hyphenation pattern has been}%
\typeout{** loaded for the language `#1'. Using the pattern for}%
\typeout{** the default language instead.}%
\else
\language=\csname l@#1\endcsname
\fi
#2}}
\providecommand{\BIBdecl}{\relax}
\BIBdecl

\bibitem{yang2024fontdiffuser}
Z.~Yang, D.~Peng, Y.~Kong, Y.~Zhang, C.~Yao, and L.~Jin, ``Fontdiffuser: One-shot font generation via denoising diffusion with multi-scale content aggregation and style contrastive learning,'' in \emph{Proceedings of the AAAI conference on artificial intelligence}, vol.~38, no.~7, 2024, pp. 6603--6611.

\bibitem{xie2021dg}
Y.~Xie, X.~Chen, L.~Sun, and Y.~Lu, ``Dg-font: Deformable generative networks for unsupervised font generation,'' in \emph{Proceedings of the IEEE/CVF conference on computer vision and pattern recognition}, 2021, pp. 5130--5140.

\bibitem{liu2019few}
M.-Y. Liu, X.~Huang, A.~Mallya, T.~Karras, T.~Aila, J.~Lehtinen, and J.~Kautz, ``Few-shot unsupervised image-to-image translation,'' in \emph{Proceedings of the IEEE/CVF international conference on computer vision}, 2019, pp. 10\,551--10\,560.

\bibitem{hassan2023real}
A.~U. Hassan, I.~Memon, and J.~Choi, ``Real-time high quality font generation with conditional font gan,'' \emph{Expert Systems with Applications}, vol. 213, p. 118907, 2023.

\bibitem{muhammad4370420learning}
A.~U.~H. Muhammad, I.~Memon, and J.~Choi, ``Learning font-style space using style-guided discriminator for few-shot font generation,'' \emph{Available at SSRN 4370420}.

\bibitem{dhariwal2021diffusion}
P.~Dhariwal and A.~Nichol, ``Diffusion models beat gans on image synthesis,'' \emph{Advances in neural information processing systems}, vol.~34, pp. 8780--8794, 2021.

\bibitem{ho2020denoising}
J.~Ho, A.~Jain, and P.~Abbeel, ``Denoising diffusion probabilistic models,'' \emph{Advances in neural information processing systems}, vol.~33, pp. 6840--6851, 2020.

\bibitem{choi2021conditioning}
J.~Choi, S.~Kim, Y.~Jeong, Y.~Gwon, and S.~Yoon, ``Conditioning method for denoising diffusion probabilistic models,'' \emph{DOI: https://doi. org/10.1109/iccv48922}, 2021.

\bibitem{saharia2022palette}
C.~Saharia, W.~Chan, H.~Chang, C.~Lee, J.~Ho, T.~Salimans, D.~Fleet, and M.~Norouzi, ``Palette: Image-to-image diffusion models,'' in \emph{ACM SIGGRAPH 2022 conference proceedings}, 2022, pp. 1--10.

\bibitem{sohl2015deep}
J.~Sohl-Dickstein, E.~Weiss, N.~Maheswaranathan, and S.~Ganguli, ``Deep unsupervised learning using nonequilibrium thermodynamics,'' in \emph{International conference on machine learning}.\hskip 1em plus 0.5em minus 0.4em\relax PMLR, 2015, pp. 2256--2265.

\bibitem{nichol2021improved}
A.~Q. Nichol and P.~Dhariwal, ``Improved denoising diffusion probabilistic models,'' in \emph{International conference on machine learning}.\hskip 1em plus 0.5em minus 0.4em\relax PMLR, 2021, pp. 8162--8171.

\bibitem{rombach2022high}
R.~Rombach, A.~Blattmann, D.~Lorenz, P.~Esser, and B.~Ommer, ``High-resolution image synthesis with latent diffusion models,'' in \emph{Proceedings of the IEEE/CVF conference on computer vision and pattern recognition}, 2022, pp. 10\,684--10\,695.

\bibitem{he2024diff}
H.~He, X.~Chen, C.~Wang, J.~Liu, B.~Du, D.~Tao, and Q.~Yu, ``Diff-font: Diffusion model for robust one-shot font generation,'' \emph{International Journal of Computer Vision}, pp. 1--15, 2024.

\bibitem{Isola_2017_CVPR}
P.~Isola, J.-Y. Zhu, T.~Zhou, and A.~A. Efros, ``Image-to-image translation with conditional adversarial networks,'' in \emph{Proceedings of the IEEE Conference on Computer Vision and Pattern Recognition (CVPR)}, July 2017.

\bibitem{Zhu_2017_ICCV}
J.-Y. Zhu, T.~Park, P.~Isola, and A.~A. Efros, ``Unpaired image-to-image translation using cycle-consistent adversarial networks,'' in \emph{Proceedings of the IEEE International Conference on Computer Vision (ICCV)}, Oct 2017.

\end{thebibliography}
\end{document}